\documentstyle[psfig,aclap]{article}  

\title{\vspace{-0.5in}Phonological modeling for continuous speech recognition in Korean}

\author{WonIl Lee, Geunbae Lee, and Jong-Hyeok Lee\\
Department of Computer Science \& Engineering \\
Pohang University of Science and Technology \\
Pohang, 790-784, Korea \\
{\tt bdragon@platon.postech.ac.kr, \{gblee,jhlee\}@vision.postech.ac.kr} \\}

\begin{document}
\bibliographystyle{fullname}
\maketitle
\vspace{-0.5in}
\begin{abstract}
	A new scheme to represent phonological changes during continuous speech
	recognition is suggested. A phonological tag coupled with its morphological
	tag is designed to represent the conditions of Korean phonological changes.
	A pairwise language model of these morphological and phonological tags
	is implemented in Korean speech recognition system. Performance of the model
	is verified through the TDNN-based speech recognition experiments.
\end{abstract}

\section{Introduction} 
The most widely used language models in speech
recognition are word-level models, such as word-pairs and word-bigrams
\cite{lee:auto}\cite{bates:bbn}\cite{agnes:spoken}. However, these models take too
much space and need large corpus to be correctly trained.
Also they are domain dependent, so it is hard to add new vocabularies.
To cope with these problems, several category-level language models are
suggested \cite{jardino:class}\cite{yang:class}.
These models include word-category models based on the fisrt and last syllables
of the words, and models using an automatic categorization technique 
to reduce the perplexity.
The category-level language models showed a reduction in space requirements
and better domain independance.
For the agglunative languages, several  morpheme category/tag-level models are
also suggested \cite{sakai:morph}\cite{nakata:morph}. These models are basically the same as
the ones used in text tagging systems,
and use bigram/trigram statistics between tags.

However, Korean has many phonological changes which happen in a
morpheme and between morphemes, and those changes result in
the disparity between phonetic and orthographic descriptions
of the morphemes. To cope with the phonological changes during 
Korean speech recognition, we suggest a representation
scheme for the phonological changes, and a morphological and phonological
tag pair language model (we call it pairwise language model).
A hierarchical morphological tag set derived from the one used in written
text analysis \cite{lee:hier} is used and a phonological tag set is 
constructed from the Korean standard pronounciation rules \cite{korean:ortho}.
Performance of the model is tested through an experimental 
TDNN(time-delayed neural network) speech recognition system.
The proposed model is quite extensible to new vocabularies and new domains
by adding new dictionary entries for the necessary morphemes, and can be refined
to bigram or trigram probabilistic models to give better recognition results.

\section{Declaritive modeling of Korean phonological rules}

Phonological changes during speech recognition in Korean are modeled
with phoneme-sequence-to-morpheme dictionary entries and
a binary connectivity matrix.

\begin{description}
\item[Phoneme-sequence-to-morpheme dictionary] 
	Figure \ref{dict0} shows a sample entry of the
	phoneme-sequence-to-morpheme dictionary. For a phoneme sequence
	[n u n], two morphemes are in the dictionary: one is
	an adnominal verb-ending and the other is a noun-ending.
	The figure shows a left and right morphological tag, and a left
	and right phonological tag for the adnominal verb-ending case.
	\footnote{A phoneme sequence can be a sequence of morphemes due to
	contraction, especially
	in spontaneous dialogues, and can have
	different left and right morphological categories.}

	The morphological tag "eCNMG" says that the morpheme is a
	verb-ending(e), makes a complex sentence(C), especially
	a inner sentence(N), through a noun phrase construction(M) and
	it is an adnominal verb-ending(G). The phonological tag "P-n" says
	that the morpheme is not changed at all(-) and the first(and the 
	last for the right phonological tag) phoneme is [n]. "P" in "P-n"
	says that it is a phonological tag.
	A phonological tag of the form "Pa=b"(see figure 2 and 4) means
	that 'a' is pronounced as [b] and "Pa2b"(see figure 3 and 4) means
	'a' is pronounced as [b] by the neutralization phenomenon.

	\begin{figure}
	\centerline{\psfig{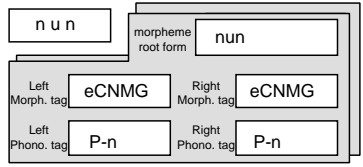}}
	\caption{Sample dictionary entry for [nun]} 
	\label{dict0}     
	\end{figure}

\item[Binary connectivity matrix]

	While the dictionary keeps the information about how a single
	morpheme is changed phonologically, the phonological binary
	connectivity matrix keeps the collocational information of
	two morphemes' pronounciations.

	Figure~\ref{matrix} shows the connectivity matrix entries for
	consonant assimilation phenomenon in Korean. These entries say that
	a morpheme whose last consonant 't s ss'\footnote{We will use 't s ss'
	notation to mean 't', 's', or 'ss' through out in this paper.} are
	changed into [n], can be followed by a morpheme whose first phoneme is
	[n] or [m]. Wild characters(*, ?) can be used to reduce the number of
	entries in the matrix.

	\begin{figure}
	\centerline{\psfig{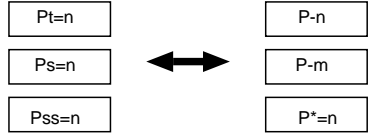}}
	\caption{Sample phonological connectivity matrix for consonant assimilation} 
	\label{matrix}    
	\end{figure}
\end{description}

Generally, we apply the following two guidelines for the phonological rule
modeling.
\begin{itemize}
\item Make a new dictionary entry for each morphologically conditioned
	phonological changes: Some phonological changes, such as vowel
	contraction and neutralization, happen only in the specific morphemes
	in a specific collocational relation.
	In these cases, registering all the phonologically changed
	morphemes or morpheme-sequences is prefered.

\item Represent the final changes when more than one changes occur:
	When more than one phonological changes occur for a morpheme or
	between morphemes, register only the final form of each morpheme
	rather than all the intermediate forms. This strategy increases
	the number of dictionary entries but eliminates the successive
	rule application.
\end{itemize}

\section{Representative phonological modeling examples}

In this section, major Korean pronounciation rules (text-to-speech rules)
 \cite{korean:ortho} are explained and their modeling (for speech-to-text conversion) 
using the dictionary and the connectivity matrix is described.
\footnote{The pronounciation rules cover intra-word and inter-word phonological
changes.}

Yale romanization is adoted to represent the Korean phonemes.

\begin{description}
\item[Neutralization\footnote{or consonant cluster simplification}]
	In Korean, only 7 consonants are pronounced as syllable coda.
	This is called neutralization or consonant cluster simplification 
	and happens when the morpheme is followed
	by a pause or a consonant.

	The followings are some examples : \\
	\begin{tabular}[th]{l l l}
	"takk+ta"	& [tak-tta]		& (clean)	\\
				&				& 'kk' $\Rightarrow$ [k] \\
	"pwu-ekh"	& [pwu-ek]		& (kitchen)	\\
				&				& 'kh' $\Rightarrow$ [k] \\
	"talk+kwa"	& [tak-kkwa]	& (chicken and)	\\
				&				& 'lk' $\Rightarrow$ [k] \\
	"os+kwa"	& [ot-kkwa]		& (clothes and)	\\
				&				& 's'~ $\Rightarrow$ [t] \\
	"nelp+ko"	& [nel-kko]		& (wide and)	\\
				&				& 'lp' $\Rightarrow$ [l] \\
	"celm+ko"	& [cem-kko]		& (young and)	\\
				&				& 'lm' $\Rightarrow$ [m] \\
	\end{tabular}

	Figure \ref{normal} shows the dictionary entry for the neutralized
	"talk"(chicken) and the corresponding connectivity matrix entry.
	"PEND" is a special tag for the pause. For each "Pa2b" tag,
	a connectivity with "PEND" is added in the matrix.

	\begin{figure}
	\centerline{\psfig{figure=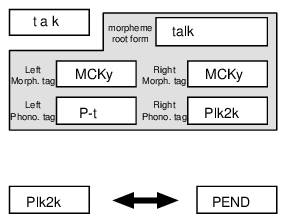}}
	\caption{Dictionary entry for [t a k] and the corresponding 
	connectivity matrix entry}
	\label{normal}    
	\end{figure}

\item[Glottalization]
	First consonants 'k t p s c' after last consonants 'k(kk, kh, ks,
	lk) t(s, ss, c, ch, th) p(ph, lp, lph, ps)' are pronounced as
	[kk tt pp ss cc] respectively.
	Verb-ending's first consonants 'k t s c' after a verb-stem with 'n(nc)
	m(lm) lp lth' as its last consonants, are pronounced as [kk tt ss cc].

	Here are some examples: \\
	\begin{tabular}[th]{l l l}
	"ppet+ta"	& [ppet-tta] & (stretch)		\\
				&			&  't' $\Rightarrow$ [tt] \\
	"iss+ten"  	& [it-tten]	 & (have existed)	\\
				&			&  't' $\Rightarrow$ [tt] \\
	"talk+kwa"	& [tak-kkwa] & (chicken and)	\\
				&			&  'k' $\Rightarrow$ [kk] \\
	"ulp+ta"	& [up-tta]	 & (recite)			\\
				&			&  't' $\Rightarrow$ [tt] \\
	\end{tabular}

	\begin{figure}
	\centerline{\psfig{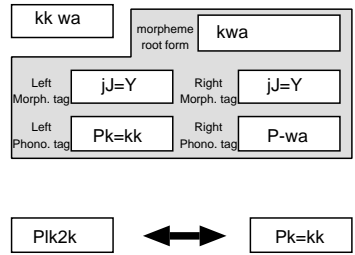}}
	\caption{Dictionary entry for [kk wa], and the corresponding 
	connectivity matrix entry}
	\label{glot}    
	\end{figure}

	Figure~\ref{glot} shows an example of "talk-kwa" which is pronounced as
	[tak-kkwa]. The connectivity matrix says that a morpheme with first
	consonant changed from 'k' into [kk] can follow a morpheme with
	last consonant 'lk' neutralized as [k].

\item[Assimilation]
	Last consonants 'k(kk, kh, ks, lk) t(s, ss, c, ch, th, h) p(ph, lp, lph)'
	followed by 'n m' are pronounced as [ng n m]. First consonant 'l'
	following the last consonants 'm ng' is pronounced as [n].
	First consonant 'l' following the last consonants 'k p' is also 
	pronounced as [n]. 'n' following or followed by 'l' is pronounced as [l].
	The first consonant 'n' following 'lh lth' is also pronounced as [l].

	\begin{tabular}[th]{l l l}
	"mek+nun"	& [meng-nun] & (eating)		\\
				&			&  'k'~ $\Rightarrow$ [ng] \\
	"iss+nun"	& [in-nun]	 & (existing)	\\
				&			&  'ss' $\Rightarrow$ [n] \\
	"tam-lyek"	& [tam-nyek] & (courage)	\\
				&			&  'l'~ $\Rightarrow$ [n] \\
	"aph+man"	& [am-man]	 & (only front)	\\
				&			&  'ph' $\Rightarrow$ [m] \\
	\end{tabular}

	\begin{figure}
	\centerline{\psfig{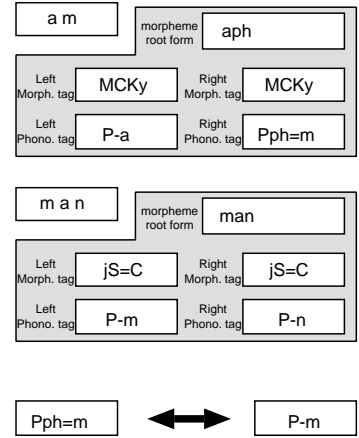}}
	\caption{Dictionary entries for [a m] and [m a n], and the corresponding
	connectivity matrix entry}
	\label{assim}    
	\end{figure}

	Figure \ref{assim} shows an example of two consecutive phonological changes
	occurred: "aph"(front) is first neutralized to [a p] and then assimilated
	to [a m]. Following the general guidelines in section 2, we model this
	phenomena as Pph=m tag. This entry in the connectivity matrix obviates
	the sequential application of several phonological rules.

\item[Consonant contraction]
	'h(nh, lh)' followed by 'k t c', are combined with those 'k t c' and
	pronounced as [kh th ch]. Final consonants 'k(lk) t p(lp) c(nc)'
	followed by 'h' are merged with that 'h' and pronounced as [kh th ph ch].
	The followings are some examples:\\

	\begin{tabular}[th]{l l l}
	"noh+ko"	 & [no-kho]		& (put down) \\
				&			&  'h'+'k' $\Rightarrow$ [kh] \\
	"manh+ko"	 & [man-kho]	& (many)	 \\
				&			&  'h'+'k' $\Rightarrow$ [kh] \\
	"talh+ci"	 & [tal-chi]	& (wear out) \\
				&			&  'h'+'c' $\Rightarrow$ [ch] \\
	"palk+hi+ta" & [pal-khi-ta] & (lighten)	 \\
				&			&  'k'+'h' $\Rightarrow$ [kh] \\
	\end{tabular}

	\begin{figure}
	\centerline{\psfig{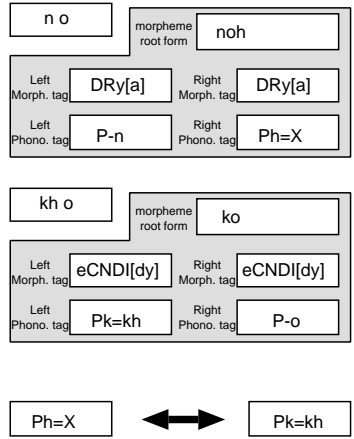}}
	\caption{Dictionary entries for [n o] and [kh o], and the corresponding
	connectivity matrix entry}
	\label{contr}    
	\end{figure}

	Figure \ref{contr} shows the case of "noh-ko". The 'h' and 'k' are merged
	to [kh]. The phonological tag "Ph=X" means that "h" is disappeared
	(changed to X(nothing)).

\item[Others]
	"cye ccye chye" in a word's conjugational form, are pronounced as
	[ce cce che].  For example, \\
	\begin{tabular}[th]{l l l l l}
	"ka-ci+e" & $\Rightarrow$ &  "ka-cye" & & \\
			& & [ka-ce] & & \\
			& & (have) & & \\
			& & 'cye'~ & $\Rightarrow$ & [ce] \\
	"cci+e" & $\Rightarrow$ & "ccye"   & & \\
			& & [cce] & & \\
			& & (cook) & & \\
			& & 'ccye' & $\Rightarrow$ & [cce] \\
	\end{tabular}

	Since the desyllabification is morphologically conditioned, the dictionary
	entries model the phenomenon according to our general guidelines. So,
	[k a c e] have the following many morphological forms in the dictionary: \\
	\begin{tabular}[th]{l l}
	[k a c e]	& "ka-ci"	\\
				& (morpheme root form) \\
				& "ka-ci+e"	\\
				& (root+sentential ending) \\
				& "ka-ci+e"	\\
				& (root+connective verb-ending) \\
				& "ka-ci+e"	\\
				& (root+aux. connective verb-ending) \\
	\end{tabular}

	However, 'yey' not in syllables "yey lyey" can be pronounced as [ey].
	In these cases, two distinct entries are made in the dictionary
	for each morpheme.

\end{description}

In this way, we modeled all the Korean pronounciation rules in about 1000
entries of phoneme-sequence-to-morpheme dictionary and more than 500 lines
of binary phonological connectivity matrix.

\section{Pair-wise language model}

The phonological connectivity matrix developed in the previous section,
coupled with the morphological connectivity matrix is used as a pair-wise
language model for continuous Korean speech recognizer.
The morphological connectivity matrix is constructed similarly to
model the Korean morphotactics using the morphological tags in the
dictionary \cite{lee:hier}.

\begin{figure}
\centerline{\psfig{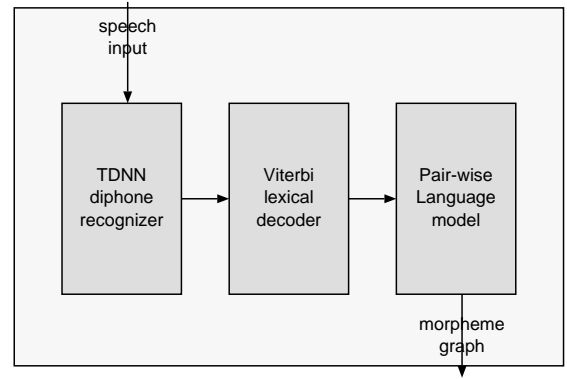}}
\caption{The TDNN-based speech recognizer} 
\label{system}    
\end{figure}

Figure \ref{system} shows the architecture of the TDNN-based continuous
speech recognizer. The TDNN-based phoneme
recognizer gvies a sequence of phoneme vectors for the input speech,
and this phoneme sequence is decoded by the Viterbi lexical decoder.
Tree-structured phoneme-sequence-to-morpheme
dictionary is used in the lexical decoding phase and a morpheme graph
is extracted after the pair-wise language model is applied.
The language model checks each adjacent pair of morphemes in the graph
whether they are connectable morphologically and phonologically.

The suggested model using the connectivity matrices for the phonological tags
and the morphological tags is easy to construct, easy to maintain,
and domain independent.
A new morpheme can be added by coding one or more dictionary entries
corresponding to its phonological variations.

\section{Experiments} 

\begin{figure}
\centerline{\psfig{figure=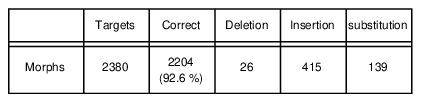}}
\caption{Morpheme recognition results for new 321 sentences}
\label{morph}    
\end{figure}

Performance of the pairwise language model is tested using the 
TDNN-based phoneme recognizer.
Input speech is sampled at 16Khz and the melscaled filter-bank output is
used as the recognizer's input.
The TDNN phoneme recognizer is trained for all 39 Korean phonemes from
the carefully selected 75 sentences (phone-balanced corpus). 
Using this recognizer, we do the Viterbi lexical decoding by employing the
tree-structured phoneme-sequence-to-morpheme dictionary,
and apply the proposed pair-wise language model.
For new 321 sentences, applying the language model produces
92.6\% correct morphemes under the
70\% correct phoneme recognition performance (figure~\ref{morph}).
The evaluation is based on the  DP best matching of the morpheme
graphs with the correct morpheme sequences.

\section{Conclusion}
In this paper, a new scheme to represent phonological changes in
Korean is suggested.  A pair-wise language model of morphological
and phonological tags is proposed for continuous Korean speech recognition.
The proposed model has the following advantages in phonological
modeling for Korean speech recognition:
\begin{itemize}
\item domain independent,
\item easy to construct,
\item easy to maintain,
\item easy to add a new vocabulary.
\end{itemize}
The pairwise language model integrates speech recognition and natural
language processing at the morpheme-level, and the morpheme-level
integration provides the full-fledged morphological/phonological 
processing which is essencial for agglunative and morphologically
complex languages, such as Korean and Japanese.
The model can be extended to categorial bigram models which are widely
used in Korean text tagging  systems.

\section{Acknowledgements}
This paper was partially supported by KOSEF (\#941-0900-084-2) and
Ministry of Information and Telecommunication, Information super-highway
application project (\#95-122).

\end{document}